\documentclass[11pt,a4paper]{article}
\usepackage[hyperref]{acl2020}
\usepackage{times}
\usepackage{float}
\usepackage{latexsym}
\usepackage{enumitem}
\usepackage{multirow}
\usepackage{multicol}
\usepackage{booktabs}
\usepackage{graphicx}
\definecolor{hulkgreen}{rgb}{0.33,0.42,0.18}

\usepackage{microtype}

\aclfinalcopy 
\def\aclpaperid{134} 

\title{\color{hulkgreen}\textsc{Hulk}\color{black}: An Energy Efficiency Benchmark Platform for \protect\\ Responsible Natural Language Processing}

\author{Xiyou Zhou, Zhiyu Chen, Xiaoyong Jin, William Yang Wang \\
  Department of Computer Science, University of California Santa Barbara\\
  \texttt{\{xiyou, zhiyuchen, x\_jin, william\}@cs.ucsb.edu}}

\date{}

\begin{document}
\maketitle
\begin{abstract}
Computation-intensive pretrained models have been taking the lead of many natural language processing benchmarks such as GLUE \citep{wang2018glue}. However, energy efficiency in the process of model training and inference becomes a critical bottleneck. We introduce \textsc{Hulk}, a multi-task energy efficiency benchmarking platform for responsible natural language processing. With \textsc{Hulk}, we compare pretrained models' energy efficiency from the perspectives of time and cost. Baseline benchmarking results are provided for further analysis. The fine-tuning efficiency of different pretrained models can differ a lot among different tasks and fewer parameter number does not necessarily imply better efficiency. We analyzed such phenomenon and demonstrate the method of comparing the multi-task efficiency of pretrained models. Our platform is available at \url{ https://sites.engineering.ucsb.edu/~xiyou/hulk/}.
\end{abstract}

\begin{table*}[ht]
    \centering
\begin{tabular}{lcccc}
\toprule
Model &  Hardware & Time & Cost & Params \\
\midrule
{BERT\textsubscript{BASE}} \citep{devlin2018bert}  & 4 TPU Pods   &     4 days & \$1,728 & 108M \\
{BERT\textsubscript{LARGE}} \citep{devlin2018bert} & 16 TPU Pods  &     4 days & \$6,912 & 334M \\
{XLNet\textsubscript{BASE}} \citep{yang2019xlnet}  & -- &  -- & --  &  117M \\
{XLNet\textsubscript{LARGE}} \citep{yang2019xlnet}  & 512 TPU v3 & 2.5 days  & \$61,440   &  361M   \\
{RoBERTa\textsubscript{BASE}} \citep{liu2019RoBERTa} & 1024 V100 GPUs &  1 day & \$75,203  &  125M   \\
{RoBERTa\textsubscript{LARGE}} \citep{liu2019RoBERTa} &  1024 V100 GPUs &  1 day  & \$75,203  &  356M  \\
{ALBERT\textsubscript{BASE}} \citep{lan2019albert}  & 64 TPU v3 & --  & --   &   12M  \\
{ALBERT\textsubscript{LARGE}} \citep{lan2019albert}  & -- & -- & -- & 18M   \\
{ALBERT\textsubscript{XLARGE}} \citep{lan2019albert} & -- & -- & -- & 59M   \\
{ALBERT\textsubscript{XXLARGE}} \citep{lan2019albert}  & 1024 TPU v3 & 32 hours  & \$65,536  & 223M  \\

{DistilBERT*} \citep{sanh2019distilbert}  &   8$\times$16G V100 GPU & 90 hours & \$2203.2 &  66M  \\
\bottomrule
\end{tabular}
\caption{Pretraining costs of baseline models. Hardware and pretraining time are collected from original papers, with which costs are estimated with current TPU price at \$8 per hour with 4 core TPU v3 chips and V100 GPU at \$3.06 per hour. DistilBERT model is trained upon a pretrained BERT model. Parameter numbers are estimated using the pretrained models implemented in the Transformers (\url{https://github.com/huggingface/transformers}) library \citep{Wolf2019HuggingFacesTS}, shown in million.}
\label{tab:pretrained model}
\end{table*}

\section{Introduction}

Environmental concerns of machine learning research has been rising as the carbon emission of certain tasks like neural architecture search reached an exceptional ``ocean boiling'' level \citep{strubell2019energy}. Increased carbon emission has been one of the key factors to aggravate global warming \footnote{Source: \url{https://climate.nasa.gov/causes/}}. Research and development process like parameter search further increase the environment impact. When using cloud-based machines, the environment impact is strongly correlated with budget.
 
The recent emergence of leaderboards such as SQuAD \citep{rajpurkar2016squad}, GLUE \citep{wang2018glue} and SuperGLUE \citep{wang2019superglue} has greatly boosted the development of advanced models in the NLP community. Pretrained models have proven to be the key ingredient for achieving state of the art in conventional metrics. However, such models can be extremely expensive to train. For example, XLNet-Large \citep{yang2019xlnet} was trained on 512 TPU v3 chips for 500K steps, which costs around 61,440 dollars\footnote{Source: \url{https://bit.ly/301qUMo}}, let alone staggeringly large carbon emission.

Moreover, despite impressive performance gain, the fine-tuning and inference efficiency of NLP models remain under-explored. As recently mentioned in a tweet\footnote{Source: \url{https://bit.ly/2GAFBNO}}, the popular AI text adventure game \it AI Dungeon \rm has reached 100 million inferences. The energy efficiency of inference cost could be critical to both business planning and environment impact.


Previous work \citep{schwartz2019green, dodge2019show} on this topic proposed new metrics like FPO (floating point operations) and new practice to report experimental results based on computing budget. Other benchmarks like \citep{coleman2017dawnbench} and \citep{mattson2019mlperf} compares the efficiency of models on the classic reading comprehension task SQuAD and machine translation tasks. However, there has not been a concrete or practical reference for accurate estimation on NLP model pretraining, fine-tunning and inference considering multi-task energy efficiency.

Energy efficiency can be reflected in many metrics including carbon emission, electricity usage, time consumption, number of parameters and FPO as shown in \citep{schwartz2019green}. Carbon emission and electricity are intuitive measures yet either hard to track or hardware-dependent. Number of parameteres does not reflect the acutal cost for model training and inference. FPO is steady for models but cannot be directly used for cost estimation. Here in order to provide a practical reference for model selection for real applications, especially model development outside of academia, we keep track of the time consumption and acutal budget for comparison. Cloud based machines are employed for cost estimation as they are easily accessible and consistent in hardware configuration and performance. In the following sections, we would use time and cost to denote the time elapsed and the acutal budget in model pretraining / training / inference.

In most NLP pretrained model setting, there are three phases: pretraining, fine-tuning and inference. If a model is trained from scratch, we consider such model has no pretraining phase but fine-tuned from scratch. Typically pretraining takes several days and hundreds of dollars, according to Table \ref{tab:pretrained model}. Fine-tuning takes a few minutes to hours, costing a lot less than pretraining phase. Inference takes several milli-seconds to seconds, costing much less than fine-tuning phase. Meanwhile, pretraining is done before fine-tuning once for all, while fine-tuning could be performed multiple times as training data updates. Inference is expected to be called numerous times for downstream applications. Such characteristics make it an intuitive choice to separate different phases during benchmarking.

\begin{figure*}
\centering
\includegraphics[scale=0.35]{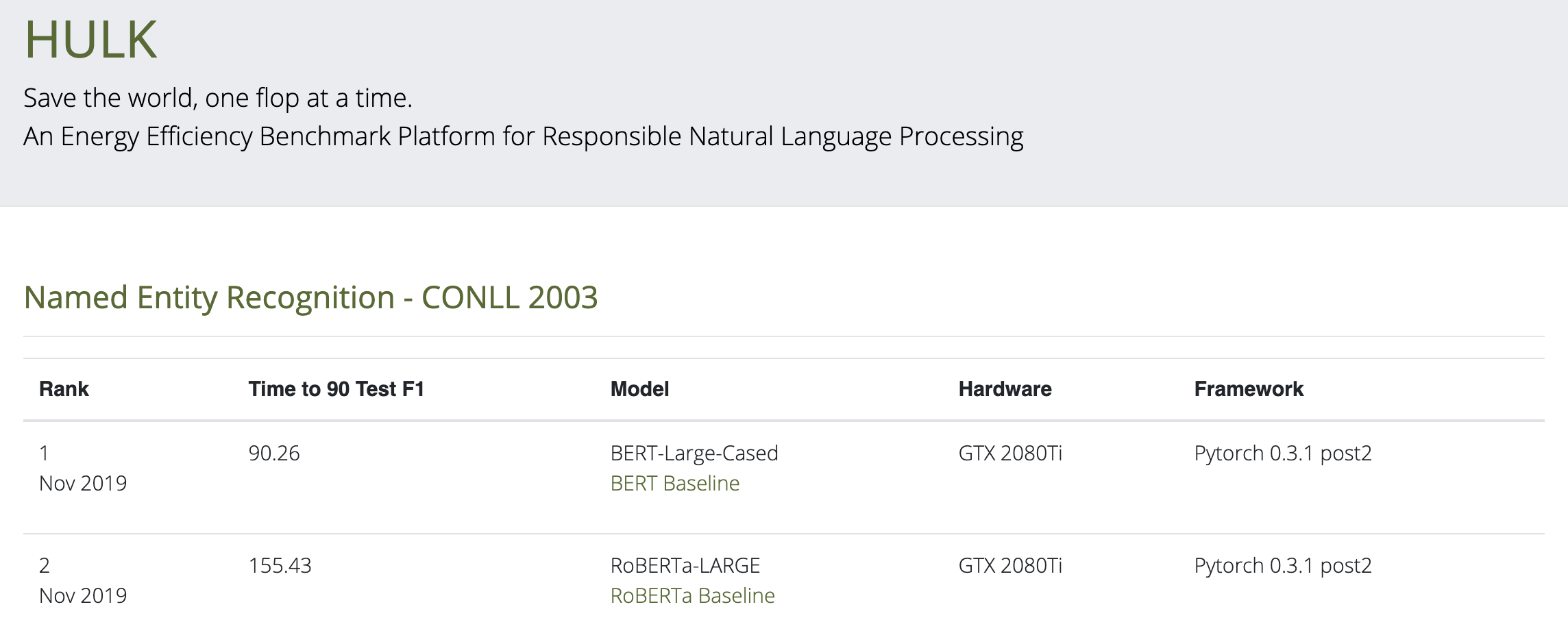}
\caption{Screenshot of the leaderboard of website.}
\label{fig_hulk}
\end{figure*}

Our \textsc{Hulk} benchmark, as shown in Figure \ref{fig_hulk}, utilizes several classic datasets that have been widely adopted in the community as benchmarking tasks to benchmark energy efficiency and compares pretrained models in a multi-task fashion. The tasks include natural language inference task MNLI \citep{williams2017broad}, sentiment analysis task SST-2 \citep{socher2013recursive} and Named Entity Recognition Task CoNLL-2003 \citep{sang2003introduction}. Such tasks are selected to provide a thourough comparison of end-to-end energy efficiency in pretraining, fine-tuning and inference.


With the \textsc{Hulk} benchmark, we quantify the energy efficiency of model pretraining, fine-tuning and inference phase by comparing the time and cost they require to reach certain overall task-specific performance level on selected datasets. The design principle and benchmarking process are detailed in section \ref{benchmarkdetails}. We also explore the relation between model parameter and fine-tuning efficiency and demonstrate consistency of energy efficiency between tasks for different pretrained models.


\section{Benchmark Overview\label{benchmarkdetails}}

For pretraining phase, the benchmark is designed to favor energy efficient models in terms of time and cost that each model takes to reach certain multi-task performance pretrained from scratch. For example, we keep track of the time and cost of a BERT model pretrained from scratch. After every thousand of pretraining steps, we clone the model for fine-tuning and see if the final performance can reach our cut-off level. When the level is reached, time and cost for pretraining is used for comparison. Models faster or cheaper to pretrain are recommended.

For fine-tuning phase, we consider the time and cost each model requires to reach certain multi-task performance fine-tuned from given pretrained models because for each single task with different difficulty and instance number, the fine-tuning characteristics may differ a lot. When pretrained models are used to deal with non-standard downstream task, especially ad hoc application in industry, the training set's difficulty cannot be accurately estimated. Therefore, it's important to compare the multi-task efficiency for model choice. 

For inference phase, the time and cost of each model making inference for single instance on multiple tasks are considered in the similar fashion as the fine-tuning phase.


\subsection{Dataset Overview}
\begin{table}
\begin{tabular}{rccc}
\toprule
           & CoNLL
           2003 & MNLI & SST-2 \\
\midrule
Train Size &  14,041    &   392,702  &    67,349 \\
Dev Size   &  3,250     &   19,647   &    872      \\
\midrule
Cut-off    &  91       &   85      &    90    \\
Metric       &    F1       &   Acc  &   Acc    \\
SOTA       &  93.5     &   91.85   &    97.4  \\
\bottomrule
\end{tabular}
\caption{Dataset Information}
\label{tab1}
\end{table}

\begin{table*}[ht]
    \centering
\begin{tabular}{lrrrrrrc}
\toprule
Datasets & \multicolumn{2}{c}{CoNLL 2003} & \multicolumn{2}{c}{SST-2} & \multicolumn{2}{c}{MNLI}\\
\midrule
Model & Time & Score & Time & Score & Time & Score & Overall Score\\
\midrule
{BERT\textsubscript{BASE}} & 43.43 & 2.08 & 207.15 &  0.45 & N/A & 0.00 & 2.53\\
{BERT\textsubscript{LARGE}}   & 90.26 & 1.00 & 92.45 &  1.00 & 9,106.72 & 1.00 & 3.00\\
{XLNet\textsubscript{BASE}} & 67.14 & 1.34 & 102.45 & 0.90 & 7,704.71 & 1.18 & 3.42\\
{XLNet\textsubscript{LARGE}}  & 243.00 & 0.37 & 367.11 & 0.25 & 939.62 & 9.69 & 10.31\\
{RoBERTa\textsubscript{BASE}} & 70.57 & 1.28 & 38.45 & 2.40 & 274.87 & 7.14 & 10.82\\
{RoBERTa\textsubscript{LARGE}}  & 155.43 & 0.58 & 57.65 & 1.60 & 397.12 & 22.93 & 25.11\\
{ALBERT\textsubscript{BASE}} & 340.64 & 0.26 & 2,767.90 & 0.03 &  N/A & 0.00 & 0.29 \\
{ALBERT\textsubscript{LARGE}}  &  844.85  & 0.11 & 3,708.49 & 0.02 &   N/A & 0.00 & 0.13\\
\bottomrule
\end{tabular}
\caption{Multi-task Baseline Fine-tuning Costs. Time is given in seconds and score is computed by the division of Time\textsubscript{BERT\textsubscript{LARGE}}/Time\textsubscript{model}.The experiments are conducted on a single GTX 2080 Ti GPU following the evaluation ceriteria. The overall score is computed by summing up scores of each individual task. For cost based leaderboads, we also use the budget to compute a new score for each task and summarize similarly. ``N/A'' means fail to reach the given performance after 5 epochs.}
\label{tab:multitask-finetune}
\end{table*}

\begin{table*}[ht]
    \centering
\begin{tabular}{lrrrrrrc}
\toprule
Datasets & \multicolumn{2}{c}{CoNLL 2003} & \multicolumn{2}{c}{SST-2} & \multicolumn{2}{c}{MNLI}\\
\midrule
Model & Time & Score & Time & Score & Time & Score & Overall Score\\
\midrule
{BERT\textsubscript{BASE}} & 2.68 & 3.18 & 2.70 &  3.13 & 2.67 & 3.19 & 9.5\\
{BERT\textsubscript{LARGE}}   & 8.51 & 1.00 & 8.46 &  1.00 & 8.53 & 1.00 & 3.00\\
{XLNet\textsubscript{BASE}} & 5.16 & 1.65 & 5.01 & 1.69 & 5.10 & 1.67 & 5.01\\
{XLNet\textsubscript{LARGE}}  & 14.84 & 0.57 & 14.69 & 0.58 & 15.27 & 0.56 & 1.71\\
{RoBERTa\textsubscript{BASE}} & 2.65 & 3.21 & 2.68 & 3.16 & 2.70 & 3.16 & 9.53\\
{RoBERTa\textsubscript{LARGE}}  & 8.35 & 1.02 & 8.36 & 1.01 & 8.70 & 0.98 & 3.01\\
{ALBERT\textsubscript{BASE}} & 2.65 & 3.21 & 2.68 & 3.18 & 2.72 & 3.14 & 9.53 \\
{ALBERT\textsubscript{LARGE}}  &  8.49  & 1.00 & 8.44 & 1.00 & 8.78 & 0.97 & 2.97\\
\bottomrule
\end{tabular}
\caption{Multi-task Baseline Inference Costs. Time is given in milliseconds and score is computed by the division of Time\textsubscript{BERT\textsubscript{LARGE}}/Time\textsubscript{model}.The experiments are conducted on a single GTX 2080 Ti GPU following the evaluation ceriteria similar to fine-tuning part. It's clear that the inference time between tasks is more consistent compared to fine-tuning phase.}
\label{tab:multitask-inference}
\end{table*}

The datasets we used are widely adopted in NLP community. Quantitative details of datasets can be found in Table \ref{tab1}.
The selected tasks are shown below:

\setlist{leftmargin=15pt,labelindent=15pt}
\setlist[enumerate]{wide=0pt, leftmargin=15pt, labelwidth=15pt, align=left}
\begin{enumerate}
    \item[] \bf CoNLL 2003 \rm The Conference on Computational Natural Language Learning (CoNLL-2003) shared task concerns language-independent named entity recognition \citep{sang2003introduction}. The task concentrates on four types of named entities: persons, locations, organizations and other miscellaneous entities. Here we only use the English dataset. The English data is a collection of news wire articles from the Reuters Corpus. Result is reflected as F1 score considering the label accuracy and recall on dev set.

    \item[] \bf MNLI \rm The Multi-Genre Natural Language Inference Corpus \citep{williams2017broad} is a crowdsourced collection of sentence pairs with textual entailment annotations. Given a premise sentence and a hypothesis sentence, the task is to predict whether the premise entails the hypothesis (entailment), contradicts the hypothesis (contradiction), or neither (neutral). The premise sentences are gathered from ten different sources, including transcribed speech, fiction, and government reports. The accuracy score is reported as the average of performance on matched and mismatched dev sets.
    
    \item[] \bf SST-2 \rm The Stanford Sentiment Treebank \citep{socher2013recursive} consists of sentences from movie reviews and human annotations of their sentiment. The task is to predict the sentiment of a given sentence. Following the setting of GLUE, we also use the two-way (positive/negative) class split, and use only sentence-level labels.
\end{enumerate}

The tasks are selected based on how representitve the dataset is. CoNLL 2003 has been a widely used dataset for named entity recognition and acutally requires output of token level labeling. NER is a core NLP task and CoNLL 2003 has been a classic dataset in this area. SST-2 and MNLI are part of the GLUE benchmark, representing sentence level labeling tasks. SST-2 has been frequently used in sentiment analysis across different generations of models. MNLI is a newly introduced large dataset for natural language inference. The training time for MNLI is relatively long and the task requires a lot more training instances. We select the three tasks for a diverse yet practical benchmark for pretrained models without constrain the models to sentence level classification tasks. In addition, their efficiency differ significantly in the fine-tuning and inference phase. Such difference can still be reflected on the final score after normalization as shown in Table \ref{tab:multitask-finetune}. Provided with more computing resource , we can bring in more datasets for even more thorough benchmarking in the furture. We illustrate the evaluation criteria in the following subsection.

\subsection{Evaluation Criteria}
In machine learning model training and inference, slight parameter change can have subtle impact on the final result. In order to make a practical reference for pretrained model selection, we compare models' end-to-end performance with respect to the pretraining time, pretraining cost, training time, training cost, inference time, infernce latency and cost following the setting of \citep{coleman2017dawnbench}.

For pretraining phase, we design the process to explore how much computing resource is required to reach certain multi-task performance by fine-tuning after the pretraining. Therefore, during model pretraining, after a number of steps, we use the half-pretrained model for fine-tuning and see if the fine-tuned model can reach our cut-off performance. When it does, we count the time and cost in the pretraining process for benchmarking and analysis.

For fine-tuning phase, we want to compare the general efficiency of pretrained model reaching cut-off performance on selected dataset.
During fine-tuning, we evaluate the half-fine-tuned model on development set after a certain number of steps. When the performance reach our cut-off performance, we count the time and cost in this fine-tuning process for benchmarking and analysis. To be specific, for a single pretrained model, the efficiency score on different tasks is defined as the sum of normalized time and cost. Here we normalize the time and cost because they vary dramatically between tasks. In order to simplify the process, we compute the ratio of BERT\textsubscript{LARGE}'s time and cost to that of each model as the normalized measure as shown in Table \ref{tab:multitask-finetune} and Table \ref{tab:multitask-inference}.

For inference phase, we follow the principles in fune-tuning except we use the time and cost of inference for benchmarking.

\subsection{Performance Cut-off Selection}
The selection of performance cutoff could be very critical because we consider certrain models being qualified after reaching certrain performance on development set. Meanwhile, certrain tasks can reach a ``sweet point'' where after relatively smaller amount of training time, the model reaches performance close to the final results despite negelagible difference. We select the cut-off performance threshold by obersvering the recent state-of-the-art performance on selected tasks.



\begin{figure}
\includegraphics[scale=0.5]{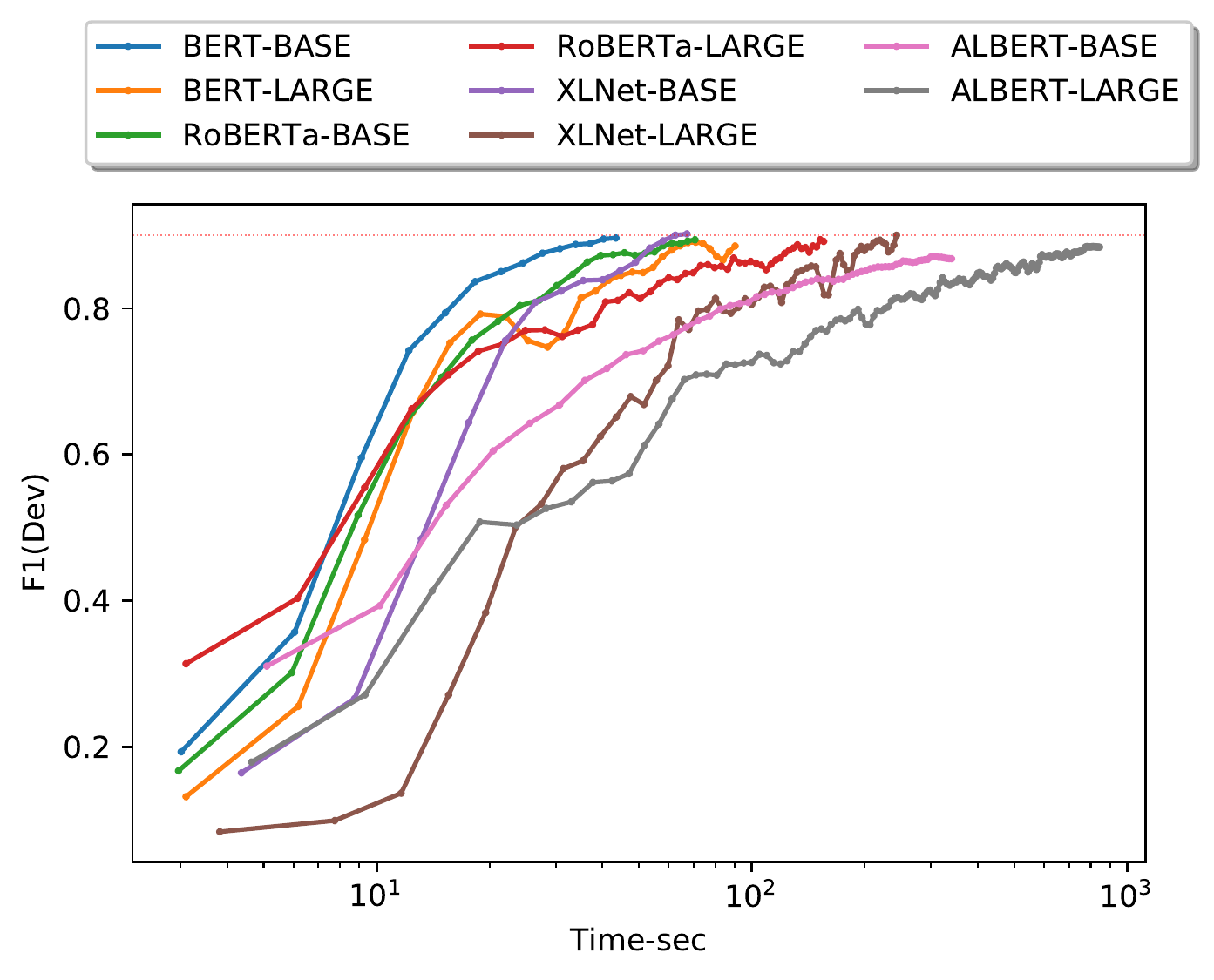}
\includegraphics[scale=0.5]{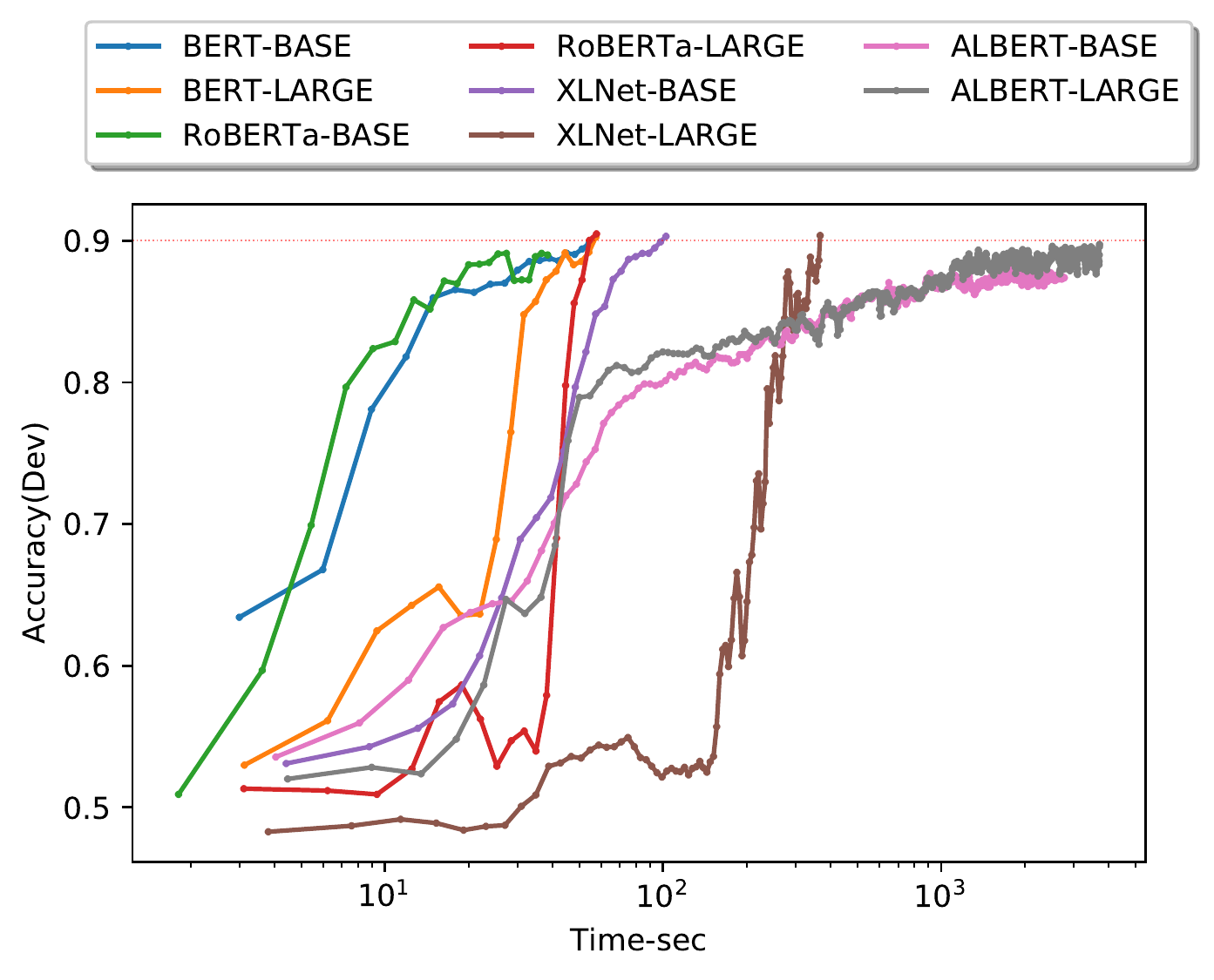}
\includegraphics[scale=0.5]{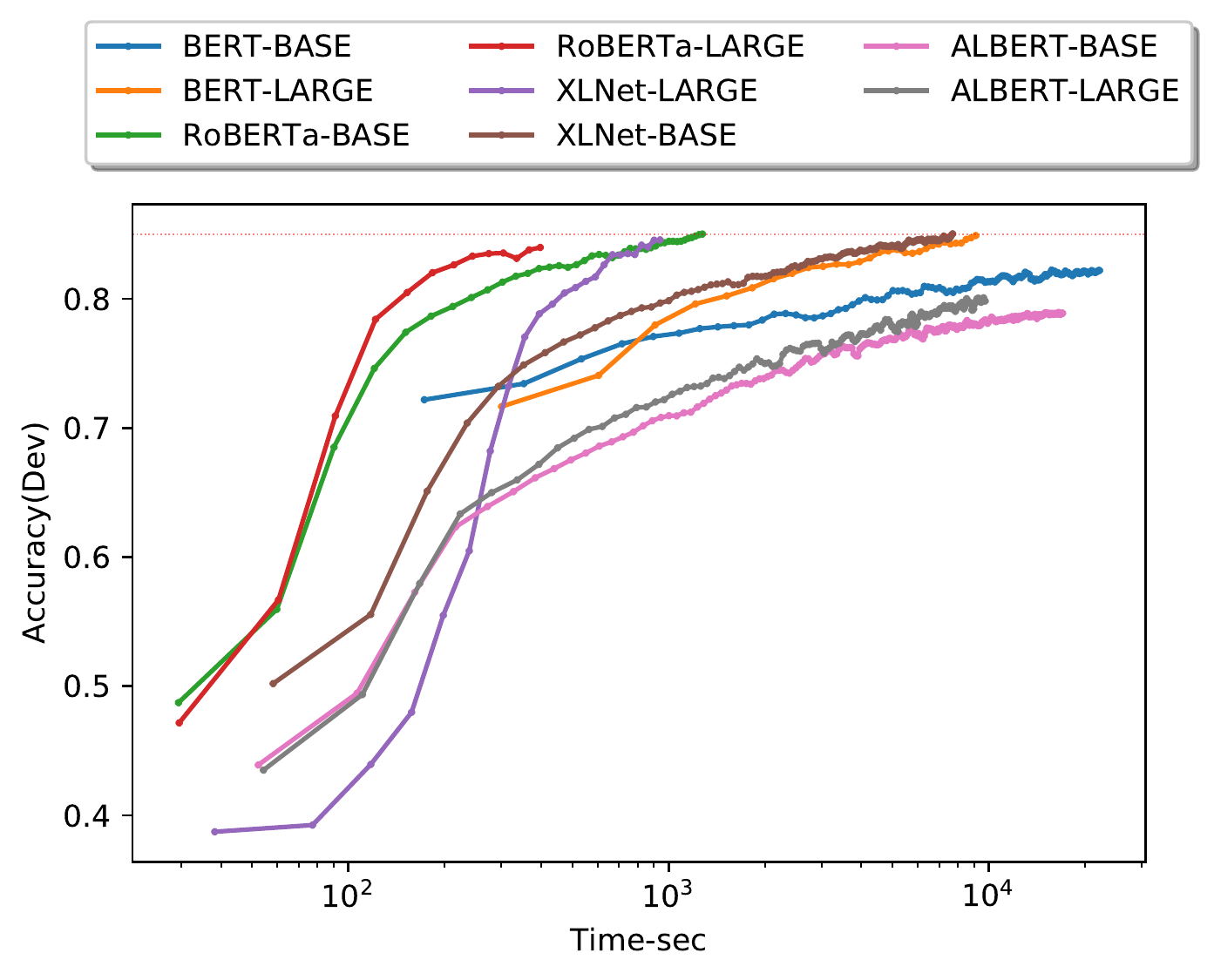}
\caption{The comparison between different pretrained models for CoNLL 2003, SST-2 and MNLI datasets trained on a single GTX 2080Ti GPU. The curves are smoothed by computing average with 2 adjacent data points. The experiments are conducted by selecting hyper-parameters to minimize the time consumption yet making sure the model can converge after certain amount of time. Results are demonstrated using performance on development score after certain steps fine-tuned on the training dataset.}
\label{fig:benchmarks}
\end{figure}

\subsection{Submission to Benchmark}
Submissions can be made to our benchmark through sending code and results to our \textsc{Hulk} benchmark CodaLab competition\footnote{The CodaLab competition is accessible from the website.} following the guidelines in both our FAQ part of website and competition introduction. We require the submissions to include detailed end-to-end model training information including model run time, cost(cloud based machine only), parameter number and part of the development set output for result validation. A training / fine-tuning log including time consumption and dev set performance after certain steps is also required. For inference, development set output, time consumption and hardware / software details should be provided. In order for model reproducity, source code is required.

\section{Baseline Settings and Analysis}

For computation-heavy tasks, we adopt the reported resource requirements in the original papers as the pretraining phase baselines.

For fine-tuning and inference phase, we conduct extensive experiments on given hardware (GTX 2080Ti GPU) with different model settings as shown in Table \ref{tab:multitask-finetune} and Table \ref{tab:multitask-inference}. We also collect the devlopment set performance with time in fine-tuning to investigate in how the model are fine-tuned for different tasks.

In our fine-tuning setting, we are given a specific hardware and software configuration, we adjust the hyper-parameter to minimize the time required for fine-tuning towards cut-off performance. For example, we choose proper batchsize and learning rate for BERT\textsubscript{BASE} to make sure the model converges  and can reach expected performance as soon as possible with parameter searching.

As shown in Figure \ref{fig:benchmarks}, the fine-tuning performance curve differs a lot among pretrained models. The x-axis denoting time consumed is shown in log-scale for better comparison of different models. None of the models acutally take the lead in all tasks. However, if two pretrained models are in the same family, such as BERT\textsubscript{BASE} and BERT\textsubscript{LARGE}, the model with smaller number of parameters tend to converge a bit faster than the other in the NER and SST-2 task. In the MNLI task, such trend does not apply possibly due to increased diffculty level and training instance number which favor larger model capacity.

Even though ALBERT model has a lot less parameters than BERT, according to Table \ref{tab:pretrained model}, the fine-tuning time of ALBERT model is significantly more than BERT models. This is probably because ALBERT uses large hidden size and more expensive matrix computation. The parameter sharing technique actually makes it harder to fine-tune the model. RoBERTa\textsubscript{LARGE} model relatively stable in all tasks.



\section{Related Work}
GLUE benchmark \citep{wang2018glue} is a popular multi-task benchmarking and diagnosis platform providing score evaluating multi-task NLP models considering multiple single task performance. SuperGLUE \citep{wang2019superglue} further develops the task and enriches the dataset used in evaluation, making the task more challenging. These multi-task benchmarks does not take computation efficiency into consideration but still innovates the development of pretrained models.

MLPerf \citep{mattson2019mlperf} compares training and inference efficiency from hardware perspective, providing helpful resources on hardware selection and model training. Their benchmark is limited to focusing on several typical applications including image classification and machine translation.

Previous work \citep{schwartz2019green, dodge2019show} on related topic working towards ``Green AI'' proposes new metrics like FPO and new principle in efficiency evaluation. We further make more detailed and practical contributions towards model energy efficiency benchmarking. 

Other work like DAWNBenchmark \citep{coleman2017dawnbench} looks into the area of end-to-end model efficiency comparison for both computer vision and NLP task SQuAD. The benchmark does not compare multi-task efficiency performance and covered only one NLP task.

The \it Efficient NMT \rm shared task of The 2nd Workshop on Neural Machine Translation and Generation proposed efficiency track to compare neural machine translation models' inference time. Our platform covers more phases and support multi-task comparison.

\section{Conclusion}
We developed the \textsc{Hulk} platform focusing on the energy efficiency evaluation of NLP models based on their end-to-end performance on selected NLP tasks. The \textsc{Hulk} platform compares models in pretraining, fine-tuning and inference phase, making it clear to follow and propose more training and inference efficient models.
We have compared the fine-tuning efficiency of given models during baseline testing and demonstrated more parameters lead to slower fine-tuning when using same model but does not hold when model changes.
We expect more submissions in the future to flourish and enrich our benchmark.

\section*{Acknowledgments}
This work is supported by the Institute of Energy Efficiency (IEE) at UCSB's seed grant in Summer 2019 to improve the energy efficiency of AI and machine learning.\footnote{\href{https://iee.ucsb.edu/news/making-ai-more-energy-efficient}{https://iee.ucsb.edu/news/making-ai-more-energy-efficient}}.

\newpage
\bibliography{acl2020}
\bibliographystyle{acl_natbib}




\end{document}